\documentclass[10pt,twocolumn,letterpaper]{article}

\usepackage{cvpr}              
\usepackage{float}

\usepackage[dvipsnames]{xcolor}

\definecolor{cvprblue}{rgb}{0.21,0.49,0.74}
\usepackage[pagebackref,breaklinks,colorlinks,citecolor=cvprblue]{hyperref}

\def\paperID{7} 
\def\confName{CVPR}
\def\confYear{2024}

\title{Investigating the Effectiveness of Cross-Attention to Unlock Zero-Shot Editing of Text-to-Video Diffusion Models}

\author{Saman Motamed$^{1}$ \and Wouter Van Gansbeke$^{1}$ \and Luc Van Gool$^{1,2,3}$ \and
\\
$^1$ INSAIT, Sofia University \quad $^2$ ETH Zurich \quad $^3$ KU Leuven}

\begin{document}
\maketitle
\begin{abstract}
With recent advances in image and video diffusion models for content creation, a plethora of techniques have been proposed for customizing their generated content. 
In particular, manipulating the cross-attention layers of Text-to-Image (T2I) diffusion models has shown great promise in controlling the shape and location of objects in the scene. Transferring image-editing techniques to the video domain, however, is extremely challenging as object motion and temporal consistency are difficult to capture accurately. In this work, we take a first look at the role of cross-attention in Text-to-Video (T2V) diffusion models for zero-shot video editing. While one-shot models have shown potential in controlling motion and camera movement, we demonstrate zero-shot control over object shape, position and movement in T2V models. We show that despite the limitations of current T2V models, cross-attention guidance can be a promising approach for editing videos. Code: \url{https://github.com/sam-motamed/Video-Editing-X-Attention.git}
\end{abstract} 
\section{Introduction}
\label{sec:intro}
\begin{figure*}
    \includegraphics[width=\linewidth]{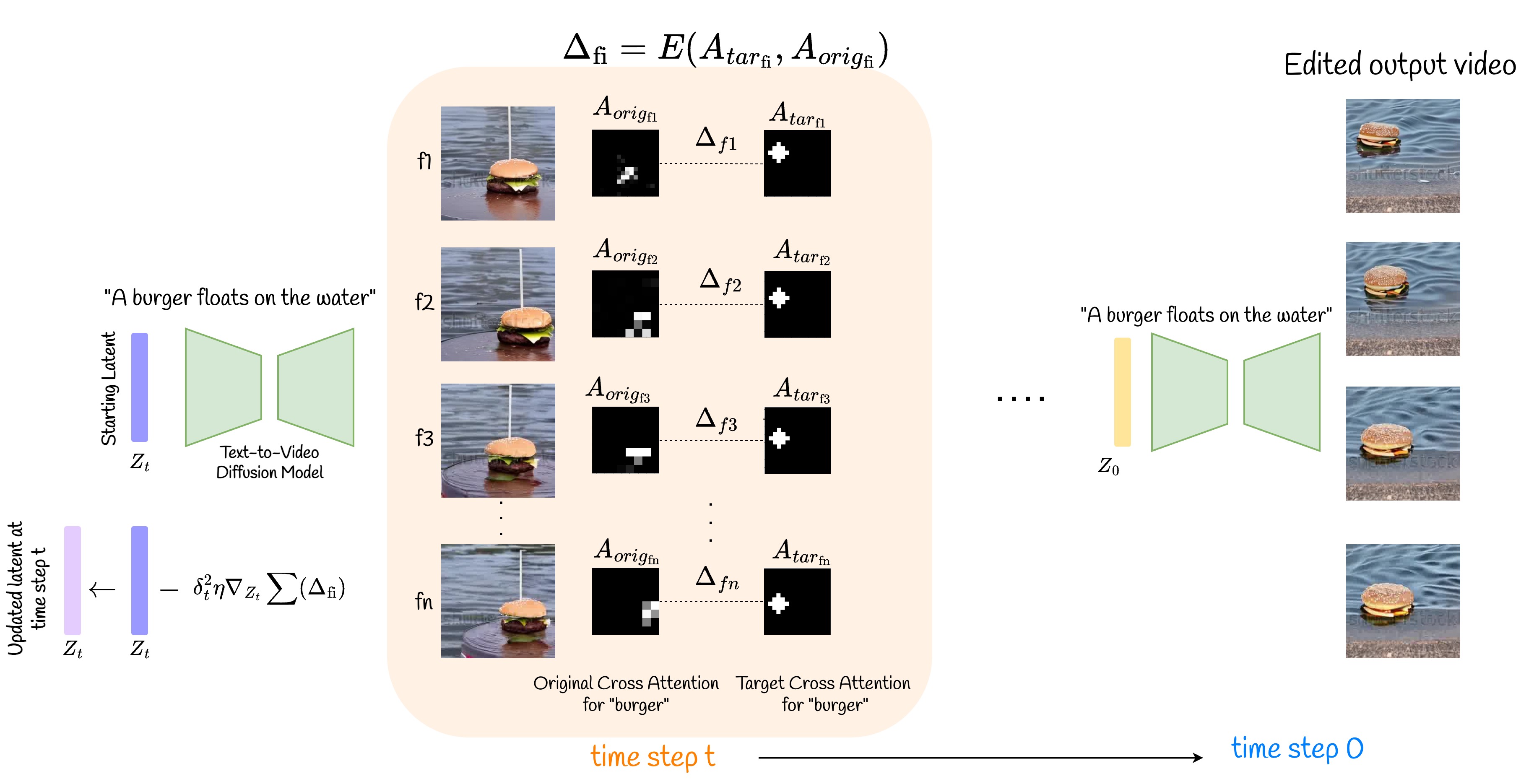}
    \caption{This figure shows an overview of backward guidance in T2V models. On the left, we show the generated frames of the T2V model after t steps, given an initial input latent $z_t$ and the text prompt ``A burger floats on the water''. To edit the video and move the burger from the top-left of the screen to the bottom-left in a straight line, we generate $\mathcal{A}_{tar_{\text{fi}}}$ for each frame \textbf{fi} reflecting this edit. Following the scheme in Section \ref{exp:details}, we update the latent through the denoising process based on objective \textbf{$E$}. At time step 0, $z_0$ generates the video on the right which reflects the intended edit.}
    \label{fig:method}
\end{figure*}
Text-to-Video diffusion models \cite{ho2022imagen,blattmann2023align,bar2024lumiere,singer2022make,wang2023modelscope} have been fast advancing in generating temporally consistent scenes with plausible object interactions. There has been a series of works that have focused on editing T2V models to enable greater control over video generation. More successful editing methods have made use of small sets of reference videos to learn an object's motion or camera movement. They subsequently transfer that specific movement or camera motion to a new object and scene \cite{zhao2023motiondirector, jeong2023vmc, yang2024directavideo} by training parts of the video diffusion model or performing Low Rank Adaptation (LoRA) \cite{hu2021lora}. While these methods can be effective, they require additional data and compute with limited flexibility, which limits their adoption in practice.

\par Several works~\cite{van2022discovering,simeoni2021localizing,wang2023cut,wang2022self} have shown the promise of attention maps in object discovery and segmentation. In the domain of text-to-image models, cross-attention and its role in controlling the scene layout has also been well studied. In particular, cross-attention is responsible for determining the objects' shape and size in the image. Cross-attention facilitates maintaining semantic consistency between the text and the generated image. By attending to relevant textual features, the model ensures that the generated visual content aligns with the overall semantics of the input description. One of the works that exploited cross-attentions to enable editing images was Prompt-to-Prompt \cite{hertz2022prompt}. This work showed that the shape of an object $\textbf{a}$ can be replaced with the shape of another object $\textbf{b}$ by replacing $\textbf{a}$'s cross-attentions with those of \textbf{b}. Training-Free Layout Control \cite{chen2023control} was another work that proposed an energy-based objective to control the position of objects in the generated image. Given a user-specified bounding box, the energy function encourages the cross-attention maps of a token to form within the bounding box and hence position the object within the bounding box. Diffusion self-guidance \cite{epstein2024diffusion} generalized the Training-Free Layout Control method such that editing the scene could be done by using the cross-attention maps alone, without the need for external inputs (e.g., bounding box), in a zero-shot manner. This was achieved by applying transformations (e.g., relocating and resizing) to the original cross-attentions of a token and using the resulting cross-attentions as the target of the objective.

\par With the success of the above methods for editing images generated by T2I models \cite{chen2024training,hertz2022prompt,epstein2024diffusion,mou2023dragondiffusion} or adapting T2I models for video editing \cite{liu2023videop2p}, we ask the question; ``Do such approaches for editing images transfer to the video domain?''. In particular, we are interested in exploring the effectiveness of cross-attention layers for editing the subject's size, positioning, and motion in videos. 

In this paper, we build upon the achievements of prior image-editing techniques by extending them to the video domain. More specifically, our contributions are threefold:
\begin{itemize}
    \item We take a first look at cross-attention layers in T2V diffusion models and their role in editing videos.
    \item We explore two possible ways to use cross-attentions in editing videos; namely \textit{forward} and \textit{backward} guidance.
    \item We investigate the limitations of current T2V models that hinder the capabilities of video editing methods.
\end{itemize}


\section{Related Works}
\label{sec:related-works}

\paragraph{Denoising Diffusion Models.} The denoising diffusion paradigm~\cite{sohl2015deep,ddpm,song2020denoising} emerged as a new method to generate images with high photo-realism and diversity. It has rapidly advanced text-conditioned image generation \cite{ldm, glide, Zhang_2023, saharia2022photorealistic, yu2022scaling, gu2022vector, ramesh2022hierarchical, ramesh2021zero, ho2022classifier, dhariwal2021diffusion}, which is important for gaining control over its generated content. 
Due to their versatility and representation learning capabilities \cite{jaini2023intriguing, yang2023diffusion, chen2024deconstructing}, they have also been successfully adapted for specialized tasks such as classification~\cite{clark2024text, Li_2023}, depth prediction~\cite{ke2023repurposing} and segmentation~\cite{van2024simple, chen2023generalist, karazija2023diffusion}.

\paragraph{Personalizing Image Generation.}
Personalizing \cite{Zhang_2023, motamed2023lego, ruiz2023dreambooth, gal2022image, wu2022generative, kumari2023multi}  and editing \cite{chen2024training, hertz2022prompt, epstein2024diffusion, mou2023dragondiffusion, patashnik2023localizing} T2I models has become a research focus to enable user-intuitive control for creating content with these generative models. In particular, the cross-attention layers of diffusion models have been studied for their role in determining a scene's layout and their ability to enable zero-shot editing of generated images. Similar to \cite{chen2024training}, we split cross-attention-based editing of T2I and T2V models into two categories of 1) \textbf{forward} and 2) \textbf{backward} guidance. 

\par\noindent In \textbf{forward} guidance, cross-attention manipulation occurs directly during the denoising process via a forward pass through the model. A notable example of forward guidance is Prompt-to-Prompt \cite{hertz2022prompt}, which proposes replacing the token's cross-attentions from a source prompt with those of a target prompt. Figure \ref{fig:fwd} shows one such example in the video domain where the cross-attentions of ``car'', from the source prompt ``car drives on the road'', are replaced with cross-attentions of ``truck", from the target prompt ``truck drives on the road". To enable more precise modifications to a specific source token, while preserving the overall scene, forward guidance requires source and target prompts that differ by a single token, limiting its applicability.

\par\noindent In contrast to forward guidance that directly manipulates cross-attentions, \textbf{backward} guidance biases the cross-attention through backpropagation. By designing an energy-based loss that encourages some desired edit \cite{chen2024training, epstein2024diffusion}, the gradient of the loss is then used to update the input
latent $z_t$ of the model. Training-Free Layout Control \cite{chen2024training} is an example of backward guidance where the energy function encourages the cross-attentions of the user-specified token to obtain higher values inside a user-defined bounding box. At multiple time steps, the input latent is updated to realize this objective. Similarly, Diffusion self-guidance \cite{epstein2024diffusion} designed energy functions that encourage the cross-attentions to take certain shapes or positions within the image. This paper is inspired by the success of these two works in the image domain. In Section \ref{sec:fwd}, we show that forward guidance is too restrictive to enable effective video editing. In Section \ref{sec:back}, we show backward guidance's promise in enabling zero-shot editing of T2V models.

\paragraph{Text-to-Video Generation.} Diffusion models have been improving at high-quality video generation by training conditional denoising networks (e.g, 3D U-Net \cite{cciccek20163d}, DiT \cite{peebles2023scalable}) to denoise randomly sampled sequences of Gaussian noises \cite{ho2022imagen, blattmann2023align, bar2024lumiere, singer2022make, wang2023modelscope, mei2023vidm}. Some works take advantage of large, pre-trained text-to-image foundation models to build text-to-video models. This is done by inflating the T2I model with temporal layers, like Tune-A-Video \cite{wu2023tune}, Text2Video-Zero \cite{khachatryan2023text2video} and AnimateDiff \cite{guo2023animatediff}.

\paragraph{Personalizing Video Generation.} Following the same desire to control image generation, a few works focused on video editing and customizing the motion and camera movement in T2V models \cite{zhao2023motiondirector, jeong2023vmc, yang2024directavideo, wang2024videocomposer, chen2023control}. Most current editing and customization methods work by tuning parts of the network or performing LoRA \cite{hu2021lora} based on example videos containing the desired effect. Such methods lack the flexibility of a zero-shot approach and require additional training data and resources. For this reason, we investigate the effectiveness of forward and backward guidance using cross-attention for T2V models.

\section{Method}
\label{sec:method}

\subsection{How Do Video Diffusion Models Work?}
Video diffusion models train a 3D denoising network, traditionally U-Nets but more recently transformer-based \cite{peebles2023scalable} networks, to generate videos from randomly sampled Gaussian noise. In this work, we use T2V models with 3D U-Net backbone \cite{wang2023modelscope} which consists of down-blocks, middle-blocks, and up-blocks. Each block has several convolution layers, spatial transformers, and temporal transformers. During training on videos, the U-Net ($\epsilon_{\theta}$) and a text encoder ($\tau_{\theta}$) are optimized with the following objective:

\begin{equation}
\mathcal{L} = \mathbb{E}_{z_0,y,\epsilon \sim \mathcal{N}(0,I), t \sim \mathcal{U}(0,T)} = \lVert \epsilon - \epsilon_{\theta}(\mathbf{z}_t, t, \tau_{\theta}(y)) \rVert^2_2,
\label{eq:t2vloss}
\end{equation}

\noindent where $z_0 \in \mathbb{R}^{f \times b \times h \times w \times c}$  is the initial latent input of the training videos ($b$ indicates the batch size, $f$ is the number of frames, $h$, $w$ and $c$ are the height and width and channels respectively) and $y$ is the text description of the video, with $\epsilon$ and $t$ being the added Gaussian noise to the videos and the time step. At time step $t$, the noised latent is defined as:
\begin{equation}
    z_t = \sqrt{\bar{\alpha}_t} z_0 + \sqrt{1 - \bar{\alpha}_t} \epsilon,
\end{equation}

 \noindent where ${\alpha}_t$ controls the noise strength.

\subsection{T2V Cross-attention}
 The cross-attention mechanism in the spatial transformers of the 3D U-Net enables the model to capture spatial relationships between the video frames and the input text. In this work, we focus on changing an object's size, location and motion given a latent input and text prompt to the T2V model. To this end, we work with the cross-attention layers of the 3D U-Net where $\{\mathcal{A}_{i,t,.,.,k} \in \mathbb{R}^{H_i \times W_i \times |k|}$\} is the Softmax-normalized cross-attention map of the $i^{th}$ layer of the U-Net, at time step $t$ for token $k$.

\subsection{Forward T2V Guidance}
\label{sec:fwd}
\begin{figure*}
    \includegraphics[width=\linewidth]{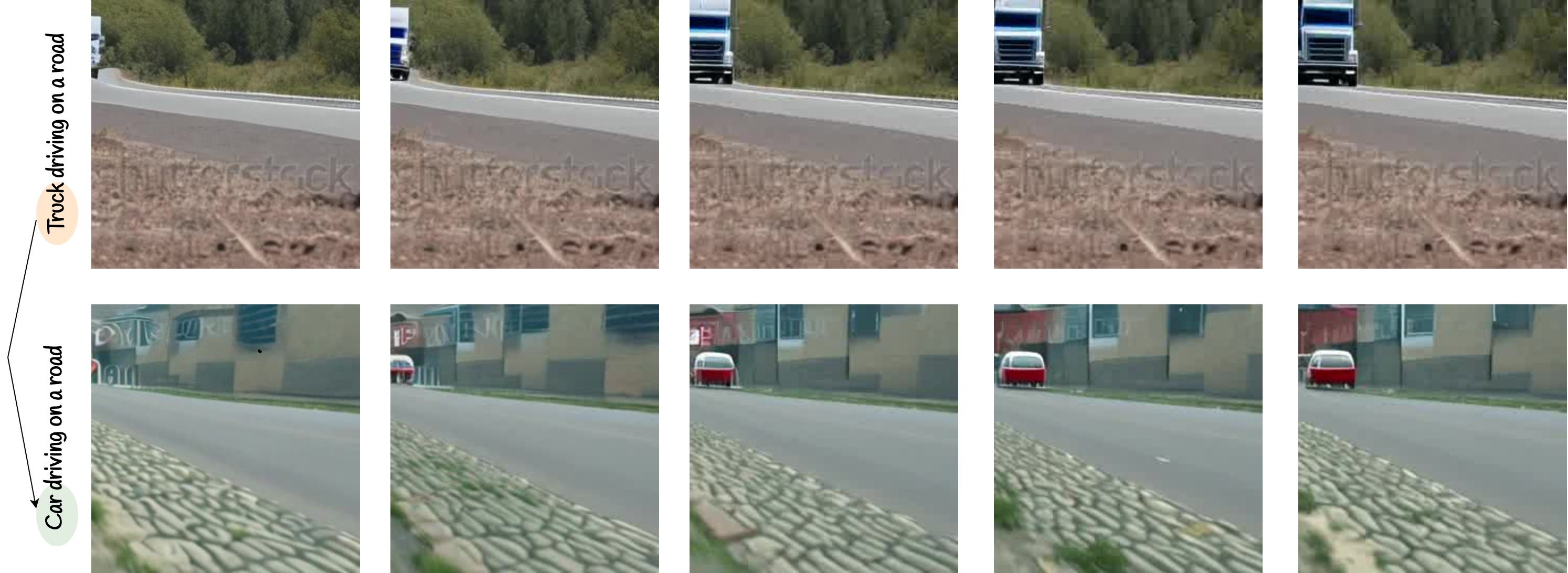}
    \caption{We show an example of forward guidance by swapping the cross-attention maps of ``car'' with cross-attention maps of the ``truck''. The two input texts only differ in one token (``truck'' and ``car''). While the car follows the motion and location of the truck in the video, artifacts can be seen around the car due to the mismatch in size and shape of the truck and car.}
    \label{fig:fwd}
\end{figure*}

Following the works that perform forward guidance in T2I models \cite{hertz2022prompt, patashnik2023localizing, chefer2023attend}, we implemented forward guidance in the T2V pipeline. Figure \ref{fig:fwd} is one example where the cross-attentions of ``car" are replaced with the cross-attentions of ``truck''. Below are the two main limitations with forward guidance that have also been observed in the T2I domain.

\begin{figure*}
\centering
    \includegraphics[width=\linewidth]{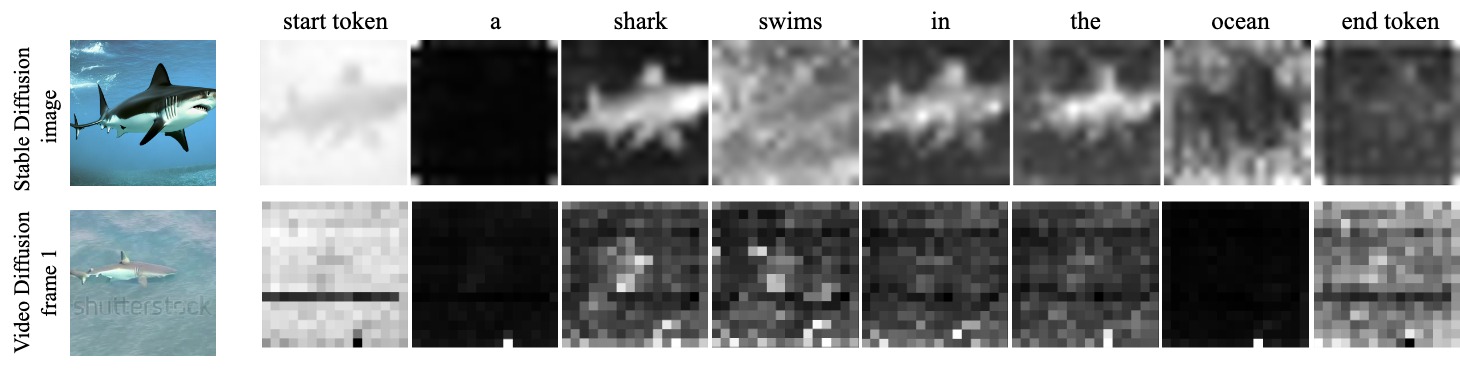}
    \caption{We compare the cross-attention maps for the same prompt to a T2I and T2V model. The cross-attention maps are extracted and averaged at the $16 \times 16$ resolution from the mid-blocks and up-blocks of the U-Net. Open-source T2I models currently produce much less noisy cross-attention maps compared to T2V models. In Section \ref{label:detailxatt}, we give details on how the noisy cross-attentions hinder backward guidance and propose a procedure for bypassing this limitation for our experiments in this paper.}
    \label{fig:xattcompare}
\end{figure*}

\begin{itemize}
    \item \textbf{Size and Shape Mismatch}. \quad Forward guidance is restrictive and can lead to artifacts due to the difference in shape and size of the two objects. In the example of Figure \ref{fig:fwd}, since the truck is larger than the car, injecting the cross-attentions of the truck to replace the car's has led to artifacts around the car without changing the car's size to match the truck's. 
    \item \textbf{Cross-attention Overlap}. \quad The cross-attentions of different tokens can overlap. We refer to the top row of Figure \ref{fig:xattcompare}, where the shark is still visible in the cross-attention maps of tokens ``in'' and ``the''. For this reason, forward guidance can work reasonably well where the two source and target sentences only differ by one token (i.e., Prompt-to-Prompt's setting). This overlap can cause degradation in the image and video quality, especially when the text inputs differ by more than one token.
\end{itemize}

We note that some of these artifacts are due to the current T2V models generating noisy cross-attentions. We go over more details in Section \ref{label:detailxatt} regarding this limitation.

\subsection{Backward T2V Guidance}
\label{sec:back}
 Following Diffusion self-guidance \cite{epstein2024diffusion} and Training-Free Layout Control \cite{chen2024training}, we define an energy function $E$ to encourage specific shape, size and motion properties on the cross-attentions of some user-specified token $k$. 
 Figure \ref{fig:method} gives an overview of our backward guidance where 
 $\mathcal{A}_{orig_{\text{fi}}}$ is the cross-attention map of some user-specified token $k$ (e.g., token corresponding to ``burger'') in frame fi of the video generated by the T2V model. we omit the layer number and the token $k$ in our notation of the cross-attention. $\mathcal{A}_{tar_{\text{fi}}}$ is the target cross-attention that captures the properties of the editing task. In Figure \ref{fig:method}, the task is to move the burger from the top-left to the bottom-left of the scene. We define the energy function $E$ below.
 To control the shape and size of an object (indicated by token $k$) through its corresponding cross-attention maps, we threshold the attention map to eliminate the effect of background noise and overlapping attention from other tokens. This is achieved by taking a soft threshold at the midpoint of the per-channel minimum and maximum values:
\[\text{shape}(k) = \mathcal{A}_{k}^{\text{threshold}}.\] Using the thresholded original cross-attention and the target cross-attention, we define the energy function $E$ as:
\begin{equation}
    E = \text{shape}(A_{tar}) - \text{shape}(A_{orig}).
\end{equation}

This objective is zero-shot since $\text{shape}(A_{tar})$ can be computed as ($M \times \text{shape}(A_{orig})$) where $M$ defines some transformation such as resizing and relocating the original attention.
At time step $t$, we update the latent $z_t$ according to the gradient of the loss defined by the energy function $E$. This is realized through the following equation:
\begin{equation}
z_t \leftarrow z_t - \delta_t^2  \eta\nabla_{Z_t}\sum E(A_{tar}, A_{orig}),
\label{eq:latentupdate}
\end{equation}

\noindent where $\eta > 0$ controls the strength of backward guidance and $\delta_t = \sqrt{(1 - \alpha_t) / \alpha_t}$. Updating the latent $z$ in this manner indirectly influences the cross-attentions. Please refer to Section \ref{exp:details} for more details on our experimental setup.
\begin{figure*}
\centering
    \includegraphics[width=\linewidth]{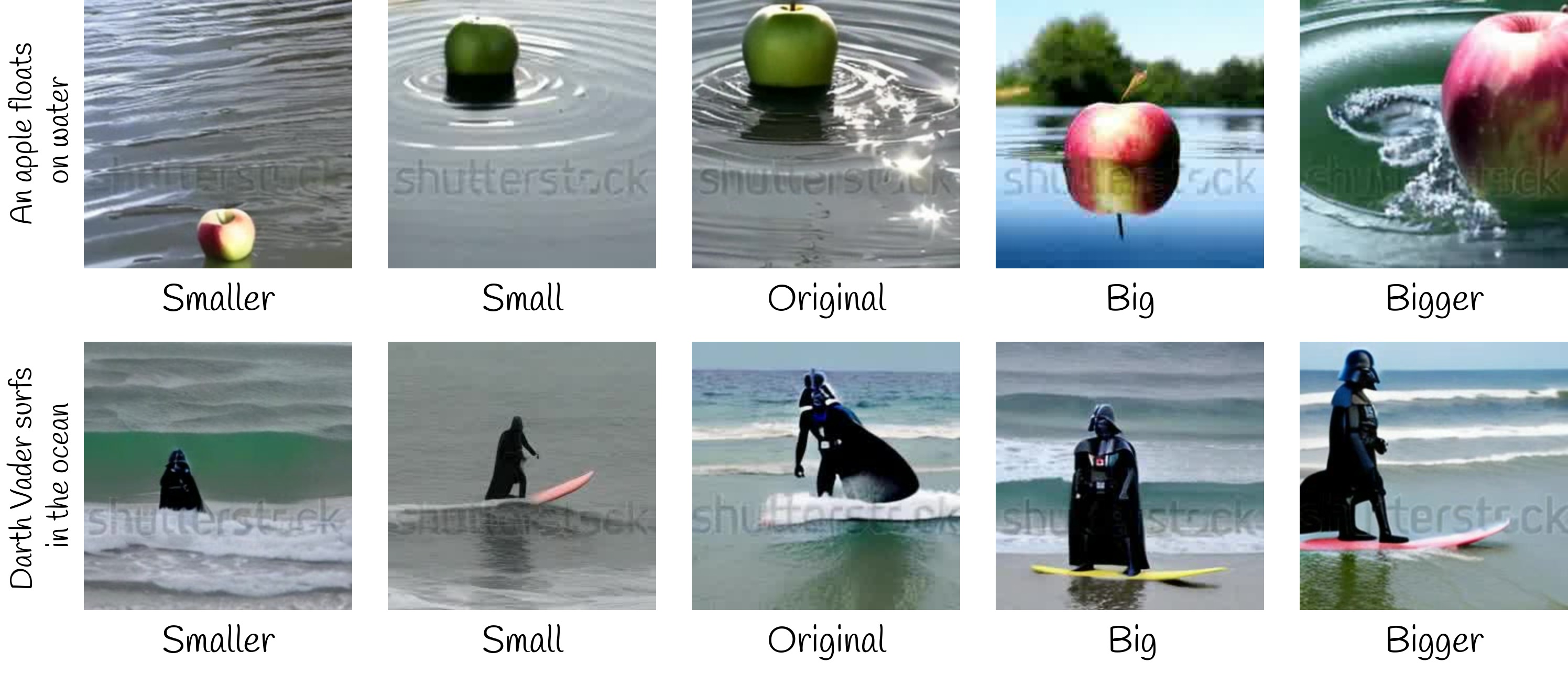}
    \caption{We show qualitative results for shrinking and enlarging objects through backward guidance. The middle image of each row visualizes the first frame of the original video. We enlarge and shrink the target cross-attentions at four different levels (Big / Bigger and small / smaller) and update the latent through backward guidance. The first frame for each edited video is shown.}
    \label{fig:resize}
\end{figure*}

\section{Experiments}
\label{sec:exp}

\subsection{Limitation of Current T2V Models}
\label{label:detailxatt}
In Figure \ref{fig:xattcompare}, we visualize the cross-attention maps for all tokens of the prompt ``a shark swims in the ocean'' generated with Stable Diffusion \cite{rombach2022high} and our T2V model \cite{wang2023modelscope}. The cross-attention maps in T2I models capture the tokens much better than T2V models. We attribute this to deeper denoising networks of T2I models, larger training datasets, and more cross-attention layers. Using such noisy cross-attention maps hinders both forward and backward guidance. To perform backward guidance more effectively, we opted to directly generate $\text{shape}(A_{tar})$ for each frame. Instead of transforming $\text{shape}(A_{orig})$ to calculate $\text{shape}(A_{tar})$ for each video frame, we generate binary cross-attention maps for the token of interest. Despite this backward guidance setup not being zero-shot, we rely on future T2V models with better cross-attention maps to replace this manual effort. 

\par Figure \ref{fig:method} shows an example of user-generated target cross-attentions. In this example, instead of transforming the cross-attention maps of the ``burger'' to calculate the target, we directly generate each frame's cross-attention according to our editing task. Here, the task is to move the burger from the top-left of the scene to the bottom-left in a straight line. Hence, we generate cross-attention maps for each frame. For frame 1, $A_{tar_{\text{f1}}}$ is placed at the top-left of the scene and in the following frames, the cross-attention map moves slightly down such that in the last frame, $A_{tar_{\text{f16}}}$ is placed at the bottom-left.

\subsection{Experiment Details}
\label{exp:details}
We use the ModelScope \cite{wang2023modelscope} T2V model in our experiments and generate 16 frame videos with $256 \times 256$ resolution. Image editing methods such as Diffusion self-guidance \cite{epstein2024diffusion} have used the extracted image features learned by the denoising network to preserve the background details and appearance features of the object being edited. In this work, we only focus on controlling objects' motion and size and leave background and appearance consistency for future works. We experimented with text prompts that describe a simple scene, to further control the limitation of current T2V models and get less noisy cross-attention maps for the object we want to edit. 

Our 3D U-Net has cross-attentions with resolutions $4 \times 4$, $8 \times 8$ , $16 \times 16$ and $32 \times 32$.  
We find the $8 \times 8$ and $16 \times 16$ cross-attentions to be the most important dimensions for effectively minimizing the energy function and editing the scene. There are 10 such layers in the down-blocks, mid-blocks, and up-blocks of the 3D U-Net (4 down-block, 2 mid-block, and 4 up-block layers). We found that mid-block's cross-attentions played a vital role in backward guidance. Excluding the two mid-block layers resulted in failed edits whereas excluding either all of down-block's or all of up-block's cross-attentions resulted in fewer failures.

\par\noindent We experimented with different schemes for updating the latent $z$ and found the most effective strategy to be that of Diffusion self-guidance \cite{epstein2024diffusion}. During the first $N/4$ iterations, we update $z$ at each step. For the subsequent $3N/4$ iterations, we update $z$ at every other step. The guidance scale $\eta$ (eq. \ref{eq:latentupdate}) also plays an important role in the method's effectiveness. Increasing $\eta$ too much leads to degradation in the generated frames. Selecting a very low scale does not change the latent enough for effective editing. We found that in our setting, a scale of $15 < \eta < 25$ provided a good balance between guidance strength and synthesis quality.

\section{Results}
\label{sec:results}
\begin{figure*}
\centering
    \includegraphics[width=0.77\linewidth]{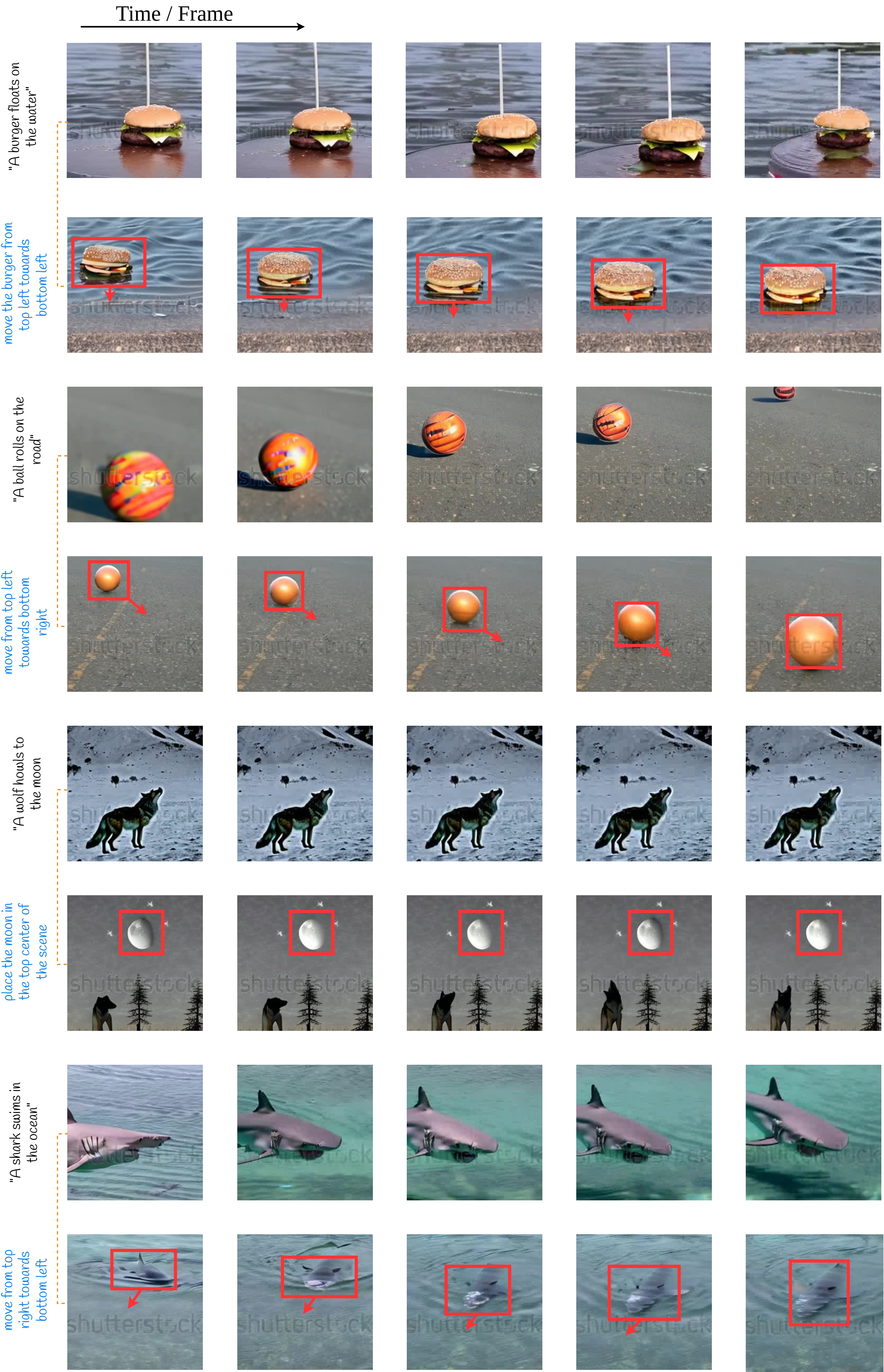}
    \caption{The figure visualizes the results of backward cross-attention guidance. For each of the 4 examples, we show the output of the T2V model given the prompt in black. The blue text describes the applied transformation to the cross-attentions at each frame. We update the input latent accordingly. The red bounding box highlights the edit's success.}
    \label{fig:result}
\end{figure*}

In this section, we show the capabilities of backward guidance for two different tasks. Figure \ref{fig:resize} shows qualitative results for changing an object's size through backward guidance. Figure \ref{fig:result} presents qualitative results of backward guidance for controlling the motion of an object in a video. To edit an object of interest, we generate binary cross-attention maps that capture the target position for the object's token. For ``burger'', we placed the first cross-attention at the top-left of the scene and slowly moved it down. For the ``ball'', we placed the cross-attention at the top-left and moved it towards the bottom-right of the scene. Finally, we moved the ``shark" from the top-right towards the bottom-left of the scene. Each sequence of frames with the black caption shows the original video without performing guidance. The sequence of frames with blue instruction shows the video after updating the latent with backward guidance. The object successfully follows the cross-attention at each frame.

We also observe that the original video can be missing an object described in the text. The example with prompt ``A wolf howls to the moon'' in Figure \ref{fig:result} is missing the moon. Interestingly, backward guidance encourages the moon to be present in the scene. Attend-and-Excite \cite{chefer2023attend} achieves the same objective in the T2I domain.

\section{Observations}
\label{observations}
In this section, we go over a few interesting observations when experimenting with backward guidance.

\paragraph{Perspective.} In our experiments, we used a fixed size for $A_{tar}$ for all 16 video frames. However, if the object is moving away or toward the camera, we should see a change in the object's size. In Figure \ref{fig:result}, we see the burger, ball, and shark getting larger as they move closer to the camera while the moon remains the same size as it is static in the sky. It is noteworthy that despite updating the model's input latent with fixed-size target cross-attentions, the model consistently generates videos with reasonable perspective. However, this comes at the expense of not strictly adhering to the exact size defined by $A_{tar}$.

\paragraph{Motion Control.} To control the motion of objects, we interpolate the cross-attention maps between the attention map $A_{tar_\text{f1}}$, placed at starting position $a$ and the attention map of the last frame $A_{tar_\text{f16}}$ placed in final position $b$. We observed that the model keeps the temporal consistency at the expense of not following the exact start and end location defined by the target cross-attention. For instance, in Figure \ref{fig:result} - last row, we placed the cross-attention of the ``shark'' at the top-right for the first frame and at the bottom-left for the last ($16^{th}$) frame. However, after $t$ steps, the shark is not at the bottom of the scene where $A_{tar_\text{f16}}$ was positioned. To do so, the model needs to move the shark much faster to go from $A_{tar_\text{f1}}$ to $A_{tar_\text{f16}}$ in a short number of frames. We also note that compared to resizing an object, controlling its motion is often prone to failures using backward guidance. This failure takes the form of the object being statically positioned at $A_{tar_\text{f1}}$.
We leave further exploration of this mode of failure for future work.

\section{Discussion}
This study conducted an initial investigation into the significance of cross-attention layers within the 3D U-Net framework of video diffusion models. More specifically, focusing on their role in determining objects' size, position and motion in T2V models. We examined the efficacy of utilizing cross-attention maps to manipulate object size and motion, employing both forward and backward guidance. In Section \ref{sec:fwd}, we showed that forward guidance in videos faces the same limitations that were previously observed in the T2I domain \cite{chen2023control} which hinders its performance. In Section \ref{sec:results}, we showed results for editing the size and motion of an object through backward guidance. Our findings emphasize the promise of backward guidance in enabling zero-shot editing capabilities for video generation. 
Moreover, in Section \ref{label:detailxatt}, we highlighted current limitations that impede the transition of cross-attention-based editing methods from the image domain to videos. This analysis provides insights into the challenges and opportunities inherent to adapting editing techniques to be used in dynamic video content.

\section{Impact and Future Directions}

Enabling zero-shot editing capabilities for generative video models is a valuable approach to enhance user control without the need for model fine-tuning with additional data. While current video models face limitations in quality, length, and cross-attention accuracy, we anticipate that editing methodologies like ours will leverage future advancements in Text-to-Video models, similar to the progress seen in the Text-to-Image domain.
\par In this study, we focused on manipulating objects' size and motion with backward guidance. However, practical applications for editing tools require further exploration, particularly enabling editing of real videos. This needs additional constraints such as controlling background alterations and maintaining the fidelity of different objects to the original video. These aspects remain open for future work.

\clearpage
{
    \small
    \bibliographystyle{ieeenat_fullname}
    \bibliography{main}
}


\end{document}